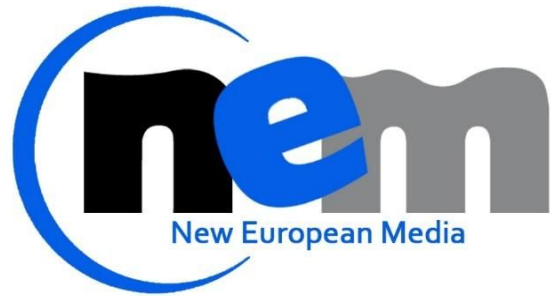

# AI in the media and creative industries

(Version 1 - April 2019)

**Editors**: Baptiste Caramiaux ([baptiste.caramiaux@inria.fr](baptiste.caramiaux@inria.fr)), Fabien Lotte ([fabien.lotte@inria.fr](fabien.lotte@inria.fr)), Joost Geurts ([jozef.geurts@inria.fr](jozef.geurts@inria.fr))

**Contributors (in alphabetical order):**

Giuseppe Amato – CNR (it),

Malte Behrmann - BBW (de),

Frédéric Bimbot – CNRS & Inria (fr),

Baptiste Caramiaux - CNRS & Inria (fr),

Fabrizio Falchi - CNR (it),

Ander Garcia - Vicomtech (es),

Joost Geurts - Inria (fr),

Jaume Gibert - Eurecat (es),

Guillaume Gravier - CNRS & Inria (fr),

Hadmut Holken - Holken consultants (fr),

Hartmut Koenitz - HKU (nl),

Sylvain Lefebvre - Inria (fr),

Antoine Liutkus - Inria (fr),

Fabien Lotte - Inria (fr),

Andrew Perkis - NTNU (no),

Rafael Redondo - Eurecat (es),

Enrico Turrin - FEP (be),

Thierry Viéville - Inria (fr),

Emmanuel Vincent - Inria (fr)

**Abstract**

Thanks to the Big Data revolution and increasing computing capacities, Artificial Intelligence (AI) has made an impressive revival over the past few years and is now omnipresent in both research and industry. The creative sectors have always been early adopters of AI technologies and this continues to be the case. As a matter of fact, recent technological developments keep pushing the boundaries of intelligent systems in creative applications: the critically acclaimed movie "Sunspring", released in 2016, was entirely written by AI technology, and the first-ever Music Album, called "Hello World", produced using AI has been released this year. Simultaneously, the exploratory nature of the creative process is raising important technical challenges for AI such as the ability for AI-powered techniques to be accurate under limited data resources, as opposed to the conventional "Big Data" approach, or the ability to process, analyse and match data from multiple modalities (text, sound, images, etc.) at the same time. The purpose of this white paper is to understand future technological advances in AI and their growing impact on creative industries. This paper addresses the following questions: Where does AI operate in creative Industries? What is its operative role? How will AI transform creative industries in the next ten years? This white paper aims to provide a realistic perspective of the scope of AI actions in creative industries, proposes a vision of how this technology could contribute to research and development works in such context, and identifies research and development challenges.

# Table of Content

# 1 Introduction: Situating modern AI in the media and creative industries

Artificial Intelligence (AI) has been undoubtedly one of the most highlighted research field in recent years. In a decade, AI has moved from an academic research topic mostly studied in departments of computer science, mathematics and psychology, to a global scientific and business incentive. AI is becoming a pervasive tool used to advance knowledge in many fields such as physics, economy, genetics, or social sciences, among others. And the technology is being deployed at the core of a wide range of applications that are used by millions of people daily.

The growing use of AI is particularly visible in the media and creative industries. As a matter of fact, creatives have always been in demand of new tools that they can use to enrich the way they work, making them early adopters of technological innovations. AI is not an exception. The technology seems to be suited to the specific requirements of the creative industries is currently profoundly changing prevailing paradigms. The purpose of this white paper is to highlight these changes and the ones expected to follow, as well as understand more in depth the methods involved behind this revolution.

What often lies behind the term "AI", at the time of writing this report, is a set of techniques able to identify complex structures from massive datasets and to use these structures to make predictions (and/or take actions and decisions) on previously unseen data. This approach is also known as Machine Learning (ML) or Statistical Learning: the computational system is able to "learn" (structures) from data and generalise to unseen data. Although AI is often used to refer to ML systems, the field encompasses a broader set of approaches such as the symbolic approach, or logic-based. In this report, we will observe that most of the current instances of AI in the media and creative industries involve ML-based systems.

The predominant use of ML, as opposed to more symbolic or rule-based approach, has to be related to the type and quantity of data at play. The gigantic daily creation, production, diffusion and consumption of texts, images, videos, and sounds on online platforms creates the substrate for the growth of AI opportunities. This means that an AI-powered system is, for example, able to tell which song one person would like to listen to (making predictions) because it has "seen" thousands of previously listened sounds by this person and "understood" the patterns representing her favourite genre, artist, song or transition between two songs. The approach is incredibly successful, thanks to the quantity of data available, the increasing computational resources, and the amazingly engineered mathematical models. And it is generic: data can be text segments (chat messages, news, articles, books, etc), images (faces, silhouettes, roads, satellite images, etc) or sounds (music, language, etc.); while predictions can be labels (names associated to faces, genre associated to music pieces), text (translation from one language to another), media (image or sound generation). AI tools can also be used to take the optimal actions or decisions, e.g., select the optimal content or difficulty to propose in an adaptive video game.

The scope of applications involving AI is growing in the media and creative industries, and this growth is fuelled by and through the rise of open-source software tools and datasets as well as low-cost computational platforms. Young companies leverage such technological solutions to create more efficiently their first prototypes. The observation holds for any practitioner wishing to embed AI in personal projects (like artists, researchers, etc.).

This report is a collaborative endeavour to gauge the extent to which AI is manifested in media and creative industries, and understand the emergent challenges in research and development. Researchers in the field of music, audio, image, videos, media and design have contributed to this report summarizing the state of the art in their own field, and collecting the outstanding challenges that remains to be solved to fully design innovative tools, knowledge and products in this sector, based on AI. We aim at identifying 1) the main creative application areas in which AI is opening promising new R&D directions in the media and creative industries, 2) AI tools that are or could be used to do so, and 3) the main scientific, technological and societal challenges that needs to solved to fully benefit from the potential of AI in the medias and creative industries.

The report is, however, not exhaustive and is not wishing to be. Indeed, the creative industry sector in itself is very broad and diverse, while AI can be applied for a rapidly growing range of purposes, making the possible ways to use AI for creative industries virtually limitless. As complementary documentation, we refer the readers to other relevant reports: the World Economic Forum's report on Creative Disruption[1], and the Villani Report on AI[2] (commissioned by the French Prime Minister).

## 2 A tour of application areas

The first aim of the report is the identify the main creative application areas in which AI is opening promising new R&D directions. We propose to structure the description related to each application around three axes: Creation, Production and Consumption/Diffusion.

**Creation** refers to the construction of new media content by any practitioner. This can be a sound designer creating new sounds for a specific scene in an action video game, or a visual artist exploring image generation until finding the image that will be used in the art work.

**Production** refers to the way media content are edited or composed in preparation to be deployed or delivered. This can be a video editor having to manually annotate video shoots before editing, or a music producer having to balance sound tracks from several musical instruments.

**Consumption** refers to the interaction an end user can have with media content. This can be someone received music recommendation on an audio streaming platform, or someone receiving news feeds on a given application on her smartphone.

### 2.1 Music

Many music-related (and audio-related) fields are currently facing important changes due to the intervention of machine learning and artificial intelligence technology in content processing. The specific challenges of audio content for machine learning relate to handling high temporal resolution and long-term structures. Early advances in machine learning for music were initially borrowed from the field of speech or language processing. Research in the field has recently become more specialized and it has exploded thanks to the creation of massive datasets from music production companies, artist-curated repositories, academic repositories and video streaming platforms. Currently, AI-based technology applied to music has gained interest in a wide range of music-related applications dispatched across creation, production and consumption.

*Creation*

The typical workflow in computer-assisted music composition is to feed the software program with scores (the input data) of a certain style or by a certain composer. The program extracts composition patterns from these scores and is able to generate new scores respecting these patterns (Briot et al., 2017; Nika et al., 2017). The very same idea is at the core of most of the so-called AI tools in music creation today: a method able to learn the underlying structure in a set of music pieces or sounds, and generates new content that sounds like the music pieces taken as examples. These tools have recently gained in complexity and expressivity, as they spread outside of academia, pushed by new incentives from the tech and music industries as well as the art world[3]. This illustrated by the fact that, while the production of the score for a musical track is often one core part of the creation process, a large body of creations are undergone through manipulating audio directly, e.g. when exploiting loops or samples.

---

[1] Creative Disruption: The impact of emerging technologies on the creative economy
https://www.weforum.org/whitepapers/creative-disruption-the-impact-of-emerging-technologies-on-the-creative-economy
[2] https://www.aiforhumanity.fr/en/
[3] An example is the Magenta group at Google: https://magenta.tensorflow.org/

A large body of research on source separation has recently enabled the *demixing* of music, allowing creatives to reuse only some particular sounds within a track, excluding the rest. Music demixing has been a major research topic for twenty years and can now be considered as solved from a scientific point of view. This calls for exploring new scientific and technological challenges which are outlined below.

Generative modeling may be considered to directly produce new musical samples after training on audio datasets. While this works well to generate temporally structured musical content and orchestration, challenges remain to be able to handle richer characteristics of sounds such as timbre. This calls for advancing the research on models able to handle better representations of audio signals. Ultimately, the goal would be to manipulate the raw audio signal directly. Some models have started to propose first solutions (see for instance (Van Den Oord et al., 2016)) but it is yet challenging to control generation of meaningful rich sounds and music through this kind of model.

Creation is fueled with inspiration, for which *style transfer* proved a very interesting technological tool in the domain of image processing, where it enabled new ways of graphical creativity[4]. In the context of music, style transfer would mean transforming an audio piece or a score so that it becomes a representative example of a target style, while retaining its specificities. For instance, transforming rock to tango, saturated to clean vocals, etc. Recent attempts have considered raw audio inputs from the classical repertoire (Mor et al., 2018). Important challenges remain: learning long-term temporal dependencies (whose scope can vary from one style to another), and allowing transfer between very different musical timbres.

In any case, these applications of AI technology to music creation are still at their infancy and can still be considered scientific challenges today. This is first due to the inherent difficulty of generating musical content, which is highly structured and requires high sampling rates, but it is also due to the difficulty of gathering large music datasets on which the systems may be trained, as opposed to the plethora of image datasets available. The main challenge is gathering symbolic music data or multitrack data that can be used as annotations of audio content. Initiatives include the Lakh MIDI dataset, which is reasonably large but which has limitations in terms of data quality, and the DSD100 multitrack dataset, but relatively small. Another dataset of significant size is AudioSet, that features musical dataset from youtube,but it is far from being an ideal resource for music research, because its core focus in on general-purpose audio processing.

AI-based music creation has also spread outside of academia. The recently released album "Hello World", advertised as the first-ever AI-based music album, involved AI as creativity-support tool, helping an artistic director to generate pieces of sound to be embedded in music soundtrack. In industry, the startup Jukedeck[5] provides musicians with a set of tools able to generate and personalized musical content. The objective is to offer new creative tools to musicians and producers as well as accelerating music making by proposing relevant elements to creatives. Another example is the London-based start-up Mogees[6] that proposes hardware-software solution for musicians to create their own musical instruments by plugging a sensor on everyday objects and by demonstrating to the system how it should sound.

### *Production*

Music production is also experiencing profound changes through the use of AI technology. The current trend for musicians is to work more and more independently from production studios, thanks to the availability of affordable technological tools. A first body of AI-based production systems then typically provide the creatives with audio engineering solutions. As an example Landr[7] is a Canadian start-up that develops solutions for mastering, distributing and communicating new music productions. As for creation, AI-based tools can be promoted as ways for musicians to independently release their music and consequently bypassing the traditional workflow of artistic direction and sound engineering. This trend is particularly supported by the wide diffusion of large-audience tools powered by well-engineered API (Application Programming Interfaces, a set of functions that can be integrated in third-party softwares and potentially commercialized). An example is the NSynth[8] tool able to generate new types of musical sounds through an underlying sound model (neural network) that has been trained offline on musical datasets. Such tool has then been used in mainstream musical industry.

---

[4] see e.g. https://www.youtube.com/watch?v=Khuj4ASldmU
[5] Jukedeck https://www.jukedeck.com/
[6] Mogees ltd. https://www.mogees.co.uk/

Music production however also remains an industry that requires professional tools. In this respect, rights holders often face the problem of repurposing legacy musical content that has a significant cultural value but a very poor audio quality: many musical standards from the 20th century are noisy, band limited and often only available in mono. There is hence a need for a new generation of tools that are able to enhance such content to make it compliant with modern audio quality standards. AI technology such as audio demixing[9] are promising tools for this purpose, providing professional sound engineers with unprecedented flexibility in audio editing. As mentioned in the previous section, music demixing has been a major research topic and recently reached maturity. However, adaptation to specific scenarios can still be challenging and require appropriately designed tools to be pragmatically used and deployed.

Although repurposing legacy content for rights holders is one key application, music creation in the studios or on stage can also strongly benefit from AI technology. In particular, much creativity is lost in the studio when musicians have to record their part independently from one another so as to reduce acoustic interference in the recorded signals. The corresponding recording time is also a waste of time and money for both the artists and the studios. A desirable feature is to process the signals originating simultaneously from all musicians, while preserving audio quality. Similarly, exploiting many low-quality sensors (such as mobile phones) that all take degraded views of an audio scene such as a concert, and combine them to reconstruct a high-quality immersive experience is an important enabling technology.

The core novelty and research challenge in the context of audio engineering is the confluence of AI and signal processing. While signal processing was mostly understood as manipulating audio samples so as to *extract* desired signals from them, AI technology now enables taking signals simply as *inputs* to sophisticated systems that can use training data so as to *extrapolate* information that has been lost and is not present in the input. This line of research is blooming in image processing[10] but is yet at its infancy for music processing.

*Consumption*

Digitization and the Internet already led to a profound change in the way music is consumed, because they enabled the end user to access virtually any music content within a few minutes. In this context, the added value for selling music moved from providing records in store to providing the users with personalized music recommendations. For this reason, recommender systems became one core activity of companies operating music streaming services such as Spotify, Deezer, Apple, Amazon, etc. Technical approaches for this purpose changed from handcrafted methods to the use of AI technology, exploiting large amounts of user session logs. Music recommender systems are now subject to a blooming research activity. However, many challenges remain and these challenges are not only technical: one major challenge in the field is the acquisition and elicitation of user preferences. While several systems rely on implicit mechanisms, this is not enough and there is a need to involve the user in the loop.

---

[7] Landr https://www.landr.com/en
[8] See for instance Sevenim's album created with Nsynth https://sevenism.bandcamp.com/album/red-blues
[9] See e.g. www.audionamix.com
[10] See e.g. https://dmitryulyanov.github.io/deep_image_prior

Another important aspect of music consumption concerns the actual *playback* technology involved. While traditional stereo systems are still omnipresent, a surge of interest in AI headphones or speakers recently appeared, where the loudspeakers are augmented with processing capabilities that enable unprecedented control over the sound such as user-specific passive noise cancelling. Similarly, mature demixing technology will soon allow the user to mute vocals from any song in real time, yielding a karaoke version in one click. New challenges can be seen in the engineering of efficient technology that can run in real-time in the wild, building upon scientific outcomes in the field.

From an even higher perspective, we may expect AI to blur the lines between music creation and music consumption, by making it possible for the user to enjoy musical content that has been specifically produced for him/her, based on past choices and user history. With the ability to demix and analyze music tracks automatically also comes the possibility to combine them so as to create new unique tracks. While musicians may produce complete songs as usual, it is likely that artists will shortly only provide some stems, to be used by automatic streaming services to generate automatic accompaniment to the taste of users.

In any case, considering existing musical content as the raw material for future music consumption also opens the path to *heritage repurposing*, where musical archives may be exploited in conjunction with more modern content to yield new and always different musical creations.

As may be envisioned, putting together AI technology and music analysis and synthesis will offer many new perspectives on the way music is consumed, that are totally in line with current trends of adding value through the analysis and the browsing of huge amounts of tracks. The next step forward appears in this sense to also add value through processing and automatic creation.

*Challenges*

- Building models able to leverage information in raw audio signals. The difficulty stems from the inherent high temporal resolution of such inputs;
- Learning long-term temporal dependencies. The difficulty is to handle various multi-scale temporal dependencies whose scope can vary from one musical style to another;
- Allowing transfer between very different musical timbres. Timbre remains not completely understood, a fully data-driven approach has yet been successful, more effort has to be done in model architecture and dataset collection to tackle this problem;
- Extrapolation information within audio signals and denoising;
- Designing efficient systems to be used by end-users for audio demixing purposes;
- Finding ways to acquire and elicitate user preferences in recommender systems.

## 2.2 Images and Visual Art

*Creation*

AI for generating art images such as photos but also non-photorealistic images is an emerging topic. Leveraging on the impressive results obtained by deep learning methods on production task such as applying for filter and style transfer, approaches have been presented to generate art.

A milestone in the direction of generating art using AI has been DeepDream [Mordvintsev 2015], a Computer Vision program created by Google. DeepDream uses a Convolutional Neural Network to find and enhance patterns in images via algorithmic pareidolia. The input image is substantially modified in order to produce desired activations in a trained deep network resulting in a dream-like hallucinogenic appearance in the deliberately over-processed images. While Deep Dream requires an image as input, the result of the process is so different from the original and so emotional for the viewer to be considered an AI generated art image.

Originally, DeepDream was designed to help to understand how neural networks work, what each layer has learned, and how these networks carry out classification tasks. In particular, instead of exactly prescribing which feature to amplify, they tested letting the network make that decision. In this case, given an arbitrary image or photo, a layer is picked and they ask the network to enhance whatever is detected. Each layer of the network deals with features at a different level of abstraction, so the complexity of generated features depends on which layer we choose to enhance. Thus, the generated image show in each part of the images what has been seen by the network in this specific part even if what has been seen is not likely to be there. Like seeing objects in clouds, the network shows what it sees even if what has been recognized is very unlikely. The dreams that can be drawn by the network are the results of the network experience. Thus, neural networks exposed during training to different images would draw different dreams even using the same image as input.

The interest of researchers on AI applied to art images has recently exploded started from 2016 when the paper "Image Style Transfer Using Convolutional Neural Networks" was presented at ECCV 2016. The proposed method used feature representations to transfer image style between arbitrary images. This paper led to a flurry of excitement and new applications, including the popular Prisma[11], Artisto[12], and Algorithmia[13]. Google has also worked on applying multiple styles to the same image.

Almost all existing generative approaches are based on Generative Adversarial Networks (GANs) [Goodfellow 2014]. The model consists of a generator that generates samples using a uniform distribution and a discriminator that discriminates between real and generated images. Originally proposed to generate images of a specific class (a specific number, person or type of object) between the ones the model has been exposed during training, GAN is now used for many other applications.

Leveraging these results, AI has also been used for generating pastiches, i.e., works of art that imitate the style of another one. In [Elgammal 2017], By building off of the GAN model, the authors built a deep-net that is capable of not only learning a distribution of the style and content components of many different pieces of art but was also able to novelly combine these components to create new pieces of art.

An interesting emerging topic is generating images from captions. Starting from the work [Mansimov 2016], various approaches have been proposed to generate images starting from captions. The goal is to generate photorealistic images, but we expect similar approaches to be applied for generating art images. The generation of high-resolution images is difficult. The higher resolution makes it easier to tell the AI generated images apart from human-generated images. However, recent works by nVidia [Karras 2018] showed exception results growing both the generator and discriminator progressively.

---

[11] https://prisma-ai.com/
[12] https://artisto.my.com/
[13] https://demos.algorithmia.com/deep-style/

Generating anime faces is the objective of [Jin 2017]. A DRAGAN-based SRResNet-like GAN model was proposed for automatic character generation to inspire experts to create new characters, and also can contribute to reducing the cost of drawing animation.

*Production*

Production can be seen as the process of creating something capitalizing on something that already exists. In the image scenario, production has various interpretations. It can be seen as the process of editing an image to produce a new one, for instance by using filters or by modifying its content. It can also be seen as the process of using existing images to produce other media, for instance using an image in a video reportage, or in a news.

Artificial intelligence has been extensively used in the image scenario with very significant results, in various applications ranging from enhancing image quality, to editing images, from image retrieval to image annotation. Most of these applications are significant for the production of images.

Artificial Intelligence was successfully applied to reproduce scene-dependent image transformations for which no reference implementation is available, as for instance photography edits of human retouchers. For instance, in [Gharbi 2017] an approach was proposed that learns to apply image transformations from a large database of input/output image transformation examples. The network is then able to reproduce these transformations, even when the formal definition of these transformation does not exist or it is not available.

Approaches were also proposed to automatically apply photo retouching operators to enhance image quality. The use of photo retouching allows photographers to significantly enhance the quality of images. However, this process is time-consuming and require advanced skills. Automatic algorithms based on artificial intelligence are able to mimic the expert's skill and to provide users with image retouching easily. In [Yan 2014] an approach that combines deep learning and hand-crafted features is proposed to perform automatic photo adjustment. In contrast to other existing approaches, this approach takes into account image content semantics, which is automatically inferred and performs adjustments that depend on the image semantics itself.

Still on the image editing side, recently deep learning based techniques were proposed that allow giving an existing image an chosen artistic style while preserving its content. For instance, it is possible to modify a picture so that it looks like a Miro painting. A significant work in this direction is given by [Gatys 2015]. Here a Deep Neural Network was proposed that creates artistic images of high perceptual quality. The system was trained to be able to separate content and style information in an image, being able to manipulate and produce artistic styles out of existing images.

Artificial intelligence was also used to produce techniques for image inpainting. Image inpainting has the objective to automatically reconstruct missing or damaged parts of an image. Applications examples are restorations of damaged painting, reconstruction of an image after deletion of objects or subjects. In all these case the aim is to modify the original image restoring or editing it so that the modifications cannot be perceived. A relevant approach in this context was proposed in [Yeh 2016]. In this paper, a Deep convolutional Generative Adversarial Network was proposed that is able to predict semantic information in the missing part and to automatically replace it with meaningful content. For instance, if an eye is missing from the image the neural network is able to correctly generate it and correctly place it.

Similarly, also AI-based techniques for image resolution enhancements were proposed. The capability of neural networks to infer semantics in an image was exploited in this scenario to accurately increase the resolution of an image with an excellent quality. In [Ledig 2016], the authors proposed a generative adversarial network also able to recover photo-realistic textures from heavily downsampled images. The proposed approach is able to infer photo-realistic natural images for 4X upscaling factor.

As we stated before, the use of existing images is often necessary for the production of other new contents. For instance, images are often used in reportages, on during the production of news. In these cases, in addition to tools for editing images, also tools to be able to identify and retrieve images relevant to the producer's needs, out of possible very large image repositories, are necessary. Also, in this case, Artificial intelligence has given a significant contribution. Image retrieval is generally performed using images as queries and searching for other images similar to the queries, or using text queries describing the wanted image content. In the first case, which is typically referred as Content-Based Image Retrieval (CBIR), we need a way to compare the query image with the images in the database and to decide which are the most relevant. In the second case, we either need to associate images with textual descriptions, or to generate visual features to be used to compare images, directly from text queries.

For several years CBIR was performed relying on hand-crafted features, that is human-designed mathematical descriptions of image content that can be compared by similarity to judge the relevance of image results to the query image. Recently, a significant step forward was obtained by training deep neural networks to extract visual descriptors from images (Deep Features), encoding significant semantic information. In this case, the high similarity between features is an indication of high semantic relationships between images. Deep features can be extracted using Deep Convolutional Neural Networks, trained to perform some recognition tasks, and using the activation of neurons in an internal layer of the network as features. This is, for instance the approach proposed in [Razavian 2014], where the authors show that performance superior to other state-of-the-art systems was obtained, with the use of deep features.

In order to use text queries to retrieve images, either textual descriptions should be associated with images, or techniques able to generate visual features from text queries are needed. In both cases, artificial intelligence has recently provided significant solutions to this problem.

Artificial Intelligence can be used to automatically analyze the content of images in order to generate annotations [Amato 2017], produce captions [Mao 2014], identify objects [Redmon 2016], recognize faces [Cao 2017], recognize relationships [Santoro 2017]. This information once extracted can be associated with images and used to serve queries.

On the other side, cross-media searching techniques are able to translate query expressed in one media to queries for another media, relying on artificial intelligence techniques. For instance, it is possible to use text to search for images or vice versa. In this case, the advantage is that an image database can be indexed once, using visual features, possibly extracted using deep convolutional neural networks. Improvement of the cross-media techniques, where the vocabulary of terms and phrases that can be translated into visual features is increased, do not require to reindex the entire database of images. Just the query-time processing tools need to be replaced. This direction is pursued in [Carrara 2016], where a neural network was trained to generate a visual representation in terms deep features extracted from the fc6 and fc7 internal layers of ImageNet, starting from a text query.

### *Diffusion and Consumption*

One of the key features needed for an effective and efficient consumption of digital images is the possibility to easily and rapidly identifying and retrieving existing content, which is relevant to one's needs. However, image content is often not described, annotated, or indexed at the required level of granularity and quality to allow quick and effective retrieval of the needed pieces of information. It is still a problem, for creative industry professionals, to easily retrieve where, for instance, a specific person is handling a specific object, in a specific place. This is generally due to the fact that metadata and descriptions, associated with digital content, do not have the required level of granularity and accuracy.

Professionals that need to retrieve, consume or reuse images, for instance, journalists, publishers, advertisers, often have to rely only on experienced archivists, with a deep knowledge of the archival content they hold, to find material of their interest. However, the amount of material generated and distributed every day makes it impossible to handle it effectively and to allow professionals to easily select and reuse the most suitable material for their needs. This happens because annotating manually, with the required level of detail, the huge volumes images produced nowadays, is extremely time-consuming and thus almost impossible to afford.

Consider, for example, the news production scenario. Every day there is an army of photographers, and journalists, around the world, that send their material to news agencies, related to some event they have witnessed, hoping that their material will be used in tv news, online magazines or newspapers. However, just a small percentage of this produced digital content will be actually published and, often, most of this audiovisual material remains buried in the news archives, unexploited because not easily discoverable.

In this respect, artificial intelligence offers effective tools to address this problem. AI-based tools for content-based image retrieval, for image annotation, image captioning, face recognition, and cross-media retrieval are nowadays available that allow effective and efficient retrieval of images according to user's needs.

Recently, deep learning techniques, as for as for instance, those based on Convolutional Neural Networks (CNN) become the state-of-the-art approach for many computer vision tasks such as image classification [Krizhevsky 2012], image retrieval and object recognition [Donahue 2013]. Convolutional Neural Networks leverage on the computing power provided by GPU architectures, to be able to learn from huge training sets. A limitation of this approach is that many large-scale training sets are built for academic purposes (for instance the ImageNet dataset), and cannot be effectively used for real-life applications.

Face recognition algorithms also benefit from the introduction of deep learning approaches. Among these, DeepFace [Taigman 2014], a deep CNN trained to classify faces using a dataset of 4 million facial images belonging to more than 4000 unique identities. More recently the VGGFace 2 dataset [Cao 2017] was released which contains 2.31 million images of 9131 subjects. A ResNet-50 Convolutional Neural Network was trained on this dataset, which is also able to determine the pose and age of persons.

*Challenges*

- Generating images from the description is still challenging even if recent works have significantly improved the state-of-the-art.
- In the last few years, the main focus of images generation using AI has been photorealistic images. While the generation of art images have been proved to be possible and relevant, it is still challenging. Instead, style transfer between images can be considered solved and only minor improvements are expected.
- Many production and consumption techniques rely on the capability of automatically understanding the content of the image. AI has significantly increased the type of objects, relationships, actions, events, etc that can be recognized, but the overall task of understanding is still challenging.
- Cross-media search, as using text queries for retrieving annotated images, is a recent promising approach to allow the retrieve of images, out of huge image databases. However, cross-media search is still challenging, especially when issues of expressiveness and scalability are also taken into account.
- Techniques for automatic reconstruction of missing or damaged parts of images work well on simple scenarios as, for instance, faces. Still, improvement is required to work satisfactorily on generic natural scenes.

Computer Vision and Pattern Recognition (CVPR '14). IEEE Computer Society, Washington, DC, USA, 1701-1708. DOI: https://doi.org/10.1109/CVPR.2014.220

## 2.3 Content and Narration

### 2.3.1 Digital storytelling

Much of the creative industry relies heavily on narration and storytelling, especially in application areas such as film, TV, games, art, media and news. Story, as we know it, remains the backbone of an experience.

Narratives have a crucial function in helping individuals and societies to make sense of the world. However, the more complex our world has become, the less effective traditional linear and static narratives seem to be at performing this organizing function and thus an informed citizenry. The representation of complex issues requires a medium that can represent vast amounts of information and engage contemporary audiences.

Digital storytelling is at the heart of the new digital media in today's creative industries and the ability to tell stories in various formats for multiple platforms is becoming increasingly important. The drive today is towards creating immersive and interactive digital stories for a diversity of services and applications, spanning from pure entertainment through edutainment and training towards digital signage and advertising. Immersive and interactive experiences are often driven by sensor input (e.g., user controllers, video images or user physiological signals - see also section 2.3.2 on this last aspect) and data, enhancing the Quality of Experience (QoE), by creating the concept of that the participants will feel a higher degree of affiliation to the content by triggering more of their senses.

In traditional storytelling, the viewer always follows a structured and logical path through the story. This is known as linear storytelling. Linear storytelling is one of the main features of our common European cultural heritage, reaching back to the traditions of ancient Greece. Europe is a very important provider and probably leader in the development of story driven content and has been for centuries. The tools for creating these digital stories have to satisfy the growing creative and media industries in Europe. Europe is the home of storytelling, the cradle of the narrative principle.

Nowadays, digital stories are not limited to one single narrative like in a traditional story structure but usually provides a framework for supporting the development of a set of principal characters and details about the immediate context of the world in which they exist. These multiple narratives are spread across delivery channels where they behave independently of one another and where interactions with audiences vary. The scattered stories should all fit cohesively together to make sense collectively or feel like they belong in the same overall story framework. Experiencing distinct stories that fit within a larger shared world is a type of seriality, which we know from our experience of watching a TV series, for example. We find a larger narrative built up one episode at a time. The larger plot is progressively built from a series of linear sequential events.

In this context, AI plays an increasing role in digital. As an example in interactive storytelling in sport, Intel and the International Olympic Committee will collaborate to offer an interactive experience of the sport games to the public through AI-based technology able to combine images from multiple perspectives[14]. The data driven storytelling open for advanced use of AI and ML and development of new digital storytelling tools. Software products can already be found on the market[15]. AI supported storytelling technologies could become a stronghold of European interactive entertainment. Therefore, not only visualisation technologies or consumption models should be considered, but also AI oriented consumer analysis led technologies regarding plots at the very edge between linear and interactive.

---

[14] https://newsroom.intel.com/news-releases/ioc-intel-announce-worldwide-top-partnership-through-2024/#gs.5l3nhq

[15] see for instance  https://narrativescience.com/Products/Quill/

*Creation*

A narrative starts as a creative process as an idea in the mind of someone. For the narrative to be usable, it has to be transcribed into a form that can be eventually turned into an objective digital form. This process requires a trans disciplinary skill set and a set of tools. The design usually requires the use of electronics, sensors, actuators and tools in the form of hardware and software. For the consumers, the final set is concerned with issues such as visualization and user interfaces. There are a lot of existing current practices to this which typically vary from business to business in the creative and media industry such as publishing, news media, broadcasting, movies, gaming etc. The hardware available for the users is also ever changing and in constant development, with the mobile devices on a strong rise and the current trend of head mounted displays and rage about mixed realities. The quest is creating immersive and interactive content the market is willing to pay for. There is often a misconception that this is new, which it is not. Some elements have been around for more than 50 years, such as the invention of VR in 1962 by Morton Heilig (The Sensorama, or «Experience Theatre»)[16]. Other similar initiatives are all based on long know principles such as transmedia storytelling (jenkins 2006). AI and in particular Machine learning are used in linear storytelling for optimisation purposes and efficiency of the use of data, for instance in searching huge archives for documentaries or improving feature narratives with content. A very recent example is the making of Apollo 11[17] where all NASA and private archives of 16,25,60 and 70mm film has been scanned and put together to recreate the moon landing in its 50 anniversary year. Also a tremendous amount of voice tracks from the ground control operators have been synchronised with the visual tracks.

In sensor based digital storytelling on the other hand, the Participants can interact with the content and make their own path through the story. This is known as non-linear storytelling. Since the objective of sensor based digital storytelling is to create more immersive narratives, it seems useful to have sensor-based digital stories trigger all of our senses as much as possible, with the aim of making participants as immersed as possible, thus making them feel as much presence in the story as possible.

Non-linear storytelling by itself already has the potential to create more immersive narratives (fully personalised if taken to the extreme). The creation challenge is in the authoring and the lack of authoring tools (managing the story „branches"). Non-linear storytelling will reach need for a trade-off between the final user-experience, and the human effort needed to create it. Adding complexity through sensor based digital storytelling without being able to scale the tools will require the use of AI to make it work beyond experiments and make it commercially viable. Indeed the use of AI in interactive immersive storytelling is conceived to increase the expressive power of the narrative [Riedl and Bulitko, 2013]. A current example of the use of AI in a storytelling tool is the Cinemachine from Unity together with the Cinecast. Cinemachine and cinecast[18] lets your machine act as the filmmaker and editor for multi camera storytelling. Especially useful in eSports, where as many as 5000 cameras are used in a story you do not know where will go. The AI component edits the input from gamers and produces a storyline for the spectators. The same can be used in AI cinematography, where content can be created in real time from visual symbolic data. The AI process helps in combining traditionally separated work flows of post production and production.

Experts and non-experts authors may have different profiles and purposes. On the one hand, expert users may be comfortable with advanced authoring environments where several aspects of the final experience could be specified and different types of experiences and contents could be generated. On the other hand, non-expert users may prefer authoring tools focused on certain types of experiences where only the main options and contents have to be specified. Thus, many authoring tools embed particular perspectives on digital storytelling and there are many good technical and aesthetic reasons for diverse approaches. Therefore, "the one" universal tool might always remain elusive. It is a current challenge to design systems and techniques allowing non-experts users authoring interactive narratives [Riedl and Bulitko, 2013].

---

[16] https://patents.google.com/patent/US3050870
[17] https://www.rollingstone.com/movies/movie-features/making-the-mind-blowing-apollo-11-doc-802297/
[18] https://create.unity3d.com/cinecast

Transmedia stories have emerged as a consensus practice where separate story elements of a larger narrative can be experienced by many different audiences via a range of technology platforms. This has been boosted by the digitalization of media and the ready distribution of content across many forms. Transmedia storytelling is a complex process and the above definition only highlights some of its important components. This is largely due to the rapidly evolving technological and media stratosphere as well as the diversity of its delivery platforms that include entertainment, advertising, marketing and social media. Since it is difficult to pinpoint one single precise definition for such a concept, it becomes useful to think of transmedia storytelling in terms of its core elements of creation, production and consumption. Each element carries many nuances and is closely related to, and influenced by, the other two. In doing so, practises as diverse as digital storytelling, interactive games, immersive media experiences and user-generated media can be studied under the umbrella of transmedias and plots in the story.

AI plays a large role in the media asset management of transmedia storytelling, especially within binning, labelling and classifying content for the various platforms and the trigger points for switching between device. Cinecast is an example of such system.

*Consumption*

Within Europe, there are many initiatives in the narrative and storytelling area. The combination of AI with questions of narration and storytelling will become even more important in multi modal interactive and virtual experiences [Cavazza 2017]. For instance, in [Guerrini 2017], they proposed an "interactive movie recombination", that uses AI to plan various alternative narratives, based on user inputs, by combining video segments intelligently according to their semantic description. In this context, technologies should play a role that allow smart plots and action sequences and their corresponding tree structures to be linked and developed for interaction. The user's information can be used to test and verify the stories and carry on. AI can also have impact on the consumer driven approach concerning characters and objects. In particular, the development of the narrative itself is important here and will be linked to audio-visual signals

In a commercial sense AI is used for personalization and segmentation which is key to the smart and immersive advertisement. Also of importance is optimizing dubbing and subtitling using machine learning to optimize the processes and automize very time consuming efforts (see also section "2.4.2 Media Access Services") .

Today, we can see a return of story- based entertainment with interactive audiobooks being only the beginning.

*Challenges*
- A vast cultural challenge between disciplines with art (writing), technology and gaming requiring bridging to be able to work on AI in a trans disciplinary way
- Inclusion of AI in the complete value chain and not only in the technical aspects but in the creative, business and ethical dimensions as well
- AI approaches to enhance automation of narration and plotting technologies in interactive entertainment specifically in the context of audio signals for the era of voice recognition-based internet, but also in the context of audio-visual signals in gaming.
- There is a lack of common understanding of immersion and interactivity which makes it difficult to find out how AI can enrich the interactive narrative dimension of the experience
- There is lack of tools for creating narrative movies using game engines.

- There is a lack of collaboration and increasing fragmentation across application domains in enhancing storytelling, such as lack of collaboration between audio and visual, broadcast and journalism.
- How to ensure the author remains in control of a story (i.e the intellectual owner of the story that gets told) that is in part generated by AI.

### 2.3.2 Content in games, movies, engineering and design

*Creation & production*

Several industries are faced with the challenge of creating large and extremely detailed numerical models of shapes and environments. This considers both the geometry of the objects as well as their surface appearance (roughness, texture, color). For instance, in the video game industry large detailed imaginary places are created, that can be freely explored by users. The movie industry faces similar challenges when virtual sets have to be designed around actors -- this is true both for animated movies, but also for motion pictures in which special effects are now ubiquitous. These environments have to follow precise requirements to produce the desired effects, in terms of aesthetics, navigability, and plausibility (immersion). In design and engineering, when modeling a part, the creative process combines goals driven by structural requirements, functional efficiency, fabrication constraints as well as aesthetics. Technologies such as additive manufacturing allow to manufacture parts with details from a few tens of microns to half a meter.

In all these fields, it has become extremely challenging to produce content using standard modeling tools (CAD/CAM software), even for the most experienced designers. First, the sheer size and level of details requirements make the task daunting: virtual environments in games and movies go from buildings to entire planets. In engineering, finding the right balance between contradictory objectives often requires to go through a tedious trial and error process, exploring for possible designs. Second, the variety of constraints to consider (navigability, structural plausibility, mechanical and structural behavior, etc.) makes standard modeling extremely difficult : the designer often has to imagine what the final behavior may be, try out her designs, and iterate to refine, either through expensive numerical simulation or by actually fabricating prototypes.

To tackle these challenges, the field of Computer Graphics has developed a rich body of work around the concept of *content synthesis*. These methods attempt to automate part of the content creation process, helping the designer in various ways: automatically filling entire regions with textures [Wei 2009] or objects [Majerowicz 2013], automatically generating detailed landscapes [Zhou 2007, Cordonnier 2017], plants and cities [Prusinkiewicz 1996, Parish 2001, Vanegas 2012], and even filling building floor prints and generating environment layouts [Merrell 2007, Merrell 2010, Ma 2014].

AI methods play a major role in simplifying the content creation process. In particular, content synthesis often results in ill-posed problems or complex, contradictory optimization objectives. In addition, finding a unique, optimal (in some sense) solution is rarely the goal. The objective is rather to produce a diversity of solutions from which the user can choose from [Sims 1991, Matusik 2005, Talton 2009, Lasram 2012, Shugrina 2015]. This latter point is particularly important in the fields of engineering and design, where *generative design* is increasingly popular: the algorithm cooperates with the user and produces a large variety of valid solutions (in the technical sense) while the user explores and suggest aesthetics. In the video game industry, algorithms that generate playable levels on-demand increase replayability while reducing costs, both in terms of content creation time and in terms of storage and network bandwidth (a game level can take up to several gigabytes of data).

While most technical objectives such as connectivity, structural requirements and geometric constraints can be formulated as objective functions, evaluating aesthetics remains subjective, cultural and personal. In this particular area, AI machine learning techniques are especially suitable. One particular methodology rooted in this trend is by-example content synthesis [Wei 2009]. These techniques produce new content that *resembles* input data. By matching features such as colors, sizes, curvatures and other geometric properties, the produced content borrows the aesthetics of the input. In this particular area, there is a significant ongoing research effort to exploit latest advances in AI, such as generative adversarial networks (GANs) [Gatys 2015]. This trend is quickly propagating to all areas of content synthesis [Mitra 2018].

With the advent of additive manufacturing, content synthesis is also increasingly important in mechanical engineering and design [Attene 2018]. Indeed, one promising way to reduce the weight of parts and achieve novel material properties is to create extremely detailed structures embedded in the volumes of the manufactured objects. This encompasses the concepts of metamaterials, architectured materials and 4D printing, with direct applications in the aerospace industry, medical domain (prosthetics) and automotive industry -- in fact, as engineers are trained in these techniques we can expect the design of *all* manufactured high-end parts to be profoundly revised in light of these possibilities. However, modeling the geometry of such parts is made difficult for very similar reasons: complexity, constraints, details and scale. AI based content synthesis methods could help produce these details. In addition to these challenges, the parts must now be optimized following complex numerical objectives (structural strength, fluid dynamics, aerodynamics, vibration absorption). This require expensive simulations -- these may be performed on a single object using large computer clusters. However the generative design process requires thousands of these simulations. A key challenge is to exploit AI and machine learning to prune the space of possibilities, and focus full-scale computations only where necessary. Overall, AI will likely play a major role in unlocking the full potential afforded by novel manufacturing processes.

### *Consumption*

Video games have become one of the main forms of entertainment and a major player in the creative industries. It is currently a market gathering more revenues than other major entertainment medias such as movies and music ([web source](#)). The video game sector has been using AI tools for decades now, in particular for designing artificial characters or opponents. However, the recent progresses in machine learning tools and associated data availability opened up the door to more personalized video game experiences. In particular, video games notably aim at providing enjoyment, pleasure and an overall positive user experience to its players (which can also contribute to more sales and/or more subscriptions to the game).

A key element that has been identified to favor enjoyment and a good gaming experience is to favor the state of Flow in the players [Chen 2007]. The Flow is a psychological state of intense focus and immersion in a task (any task), during which people lose the sense of time, perform at the best of their capacity and derive the most enjoyment. Flow can notably occur when the challenges offered to each player match their skills, so that the game is neither too easy - which would be boring - nor too hard - which would be frustrating. However, various players have various skills and seek or need different content and difficulty in a game to enjoy it. Thus, to ensure maximally enjoyable games for a maximum number of players requires personalized games, whose content and challenges adapt dynamically and automatically to each player.

AI methods could be used to design such personalized games, by 1) modelling the users, 2) providing adaptive content dynamically, based on the model of each player [Cowley 2016a], thus favoring flow and game enjoyment. We detail these points below.

First, in order for a game to adapt to each player, it will have to be able to understand this player, and thus to model it automatically from available data. In particular, a useful player model would have information about the player's skills, cognitive states (e.g., attention level), conative state (i.e., motivation) and affective states (e.g., frustration or joy), and possibly the player's traits (e.g., personality). Such states, skills and traits could possibly be inferred from various data available during play, using machine learning techniques (and in particular classification algorithms). For instance, some player's states, skills or traits could be inferred from the behaviour of the player in the game, depending on his/her actions and the context. The skills can also be inferred from the performance of the user in the game, see, .e.g., [Herbrich 2007] and [Bishop 2013] for a simple skill estimator based on machine learning used for Xbox live. Various cognitive, affective and conative states could also be estimated from physiological signals measured from the player, using sensors embedded in gamepads, in gaming devices, or using wearables. For instance, there are first laboratory results suggesting that some cognitive states such as workload, or some affective states (e.g., positive or negative emotions) could be inferred, to some extents, from rather common physiological signals such as from speech recordings, eye tracking, facial expressions (from photos/videos) [Nacke 2008] or from more advanced ones such as muscle activity (electromyography - EMG), heart rate, galvanic skin response (GSR) [Cowley 2016, Drachen 2010, Nacke 2013] or even from brain signals such as electroencephalography (EEG) [Frey 2014, Frey 2016, Mühl 2014]. So far, only a few states can be estimated from these measures, and rather unreliably and unspecifically. Modern AI/Machine learning tools hold promises to estimate a larger variety of states (including, for instance, the Flow state) in a robust way, in order to obtain rich and reliable player models.

Then, once these various states, skills and traits identified and estimated, AI could be used to model and estimate how game events, mechanics and/or difficulty levels would impact the player's enjoyment (as reflected by the player's estimated affective states). This would provide a comprehensive model able to predict how the player will behave and perceive the game depending on its states and skills, and depending on the game context. Once such a robust model of the player obtained, different AI tools could be used to provide adaptive content in the game, dynamically, depending on the player's model. Such AI tools could for instance adapt the game difficulty, change or alter the game story or scenario, trigger various events to induce various emotions, or provide new choices to the players. For instance, the player models defined above could be used to predict the possible impacts of various actions and game adaptations, in order to select suitable actions to maximize Flow and game enjoyment. Some authors have recently proposed that machine learning tools such as recommender systems could be used to select actions that have a positive impact on the gaming experience, based on the impact similar actions had on similar users in the past [Tondello 2017].

Extending such promising R&D directions would include studying the use of similar techniques to design serious games, e.g., for education. As it happens, in the field of Intelligent Tutoring Systems, similar user modelling and adaptive AI tools are used for providing personalized education. Such works also raise some crucial ethical questions. In particular, such player modelling and game adaptation aim at maintaining the player in the Flow zone and at maximizing game enjoyment. As such, there is naturally a risk to lead to game addiction, which was recently recognized as a disease by the World Health Organization (http://www.who.int/features/qa/gaming-disorder/en/). Such AI tools for gaming should thus also include "ethics by design", to also prevent gaming addictions.

*Challenges*

- Synthesizing large environments remains challenging, despite significant progress in specific domains. In particular, large environments are heterogeneous and require different types of content -- buildings, textures, natural elements -- that have to follow a higher-level organization -- navigability, type of buildings, distribution of classes, etc. We currently lack the algorithms and methodologies that would allow these different models to interoperate.
- AI approaches often require large input datasets to seed a machine learning algorithm. However, we also have to develop methods that can work from a sparse input, provided by a single designer or engineer describing what he seeks to achieve to an algorithm.
- The available formalisms often result in stochastic processes sampling random instances from a model learned from the content. However, these methodologies cannot easily take into account hard constraints, such as manufacturability, connectivity/navigability and structural requirements. This currently requires to perform either rejection sampling [Shugrina 2015] or progressive refinement during user exploration [Umetani 2015]. While these provide effective solutions, a challenge is to define methodologies that can sample instances directly enforcing the requirements.
- Content synthesis methods are often thought as purely automatic, stochastic processes, while most applications require a form of cooperation with a human designer. This typically requires methods than can interact with the user, with fast response times, precluding the use of expensive simulations in the loop. Similarly, we need methods that can assist users in choosing and navigating amongst a variety of possibilities, each offering different tradeoffs while enforcing specific constraints.
- We have to develop novel methodologies to analyze the output of content synthesis methods, in particular the average and variance in the desired properties. This is crucial, in particular, when producing small scale functional details for mechanical engineering applications.
- Design robust machine learning tools to estimate users' states (cognitive, conative and affective states) reliably, from their behaviour and physiological signals, while they interact with a game, while they watch a movie or use an educational software
- Identify what can be adapted in games and movies and how, and propose a taxonomy of that in order to increase flow and other desirable mental states during play.
- Extend and adapt the AI tools used to favor a positive user experience in game to the education sector, to favor efficient and effective learning, e.g., for serious games or other digital learning software.
- Formalize and develop "ethic by design" in AI tools favoring flow and other positive user experience, to prevent addictions.

[Umetani 2015] N. Umetani, T. Igarashi, N.J. Mitra. Guided Exploration of Physically Valid Shapes for Furniture Design. Communications of the ACM 2015

[Vanegas 2012] Inverse Design of Urban Procedural Models. C.A. Vanegas, I. Garcia-Dorado, Ignacio D. Aliaga, B. Benes, P. Waddell. ACM Transactions on Graphics 6(31), 2012

[Wei 2009] L-Y. Wei, S. Lefebvre, V. Kwatra, and G. Turk. State of the art in example-based texture synthesis. Eurographics 2009, State of the Art Report, EG-STAR, pages 93–117.

[Zhou 2007] H. Zhou, J. Sun, G. Turk, and J. M. Rehg. Terrain synthesis from digital elevation models. IEEE Transactions on Visualization and Computer Graphics, 13(4):834–848, 2007

## 2.4 Information and Media

### 2.4.1 AI and Book Publishing

*Production and consumption*

The book publishing sector is looking with interest at technological developments such as AI, and exploring its potential applications, through the machine learning and rule-based approaches, in fields such as reasoning on the basis of semantic web contents.

Possible areas for the use of Artificial Intelligence in the book publishing sector can be mainly distinguished between the ones that improve the efficiency of processes in publishing houses (including the development of better, more ´intelligent´, more target-group oriented products) on the one hand, and a better access to publishing contents on the side of the customer on the other.

Examples of the latter category are not only general virtual assistants, but also more specific applications like the ones that improve the accessibility for people with print impairments. Research and innovation for accessibility is necessary to increase the number of books made available in accessible format at the time of publication, allowing the same quality of reading experience for all. Image descriptions and automated tagging are fields in which applications of artificial intelligence have high innovation potential: they can help in creating alternative texts for the images in books, for example. In order to provide meaningful description for the visually impaired, it is necessary to convey the concept expressed by the images, rather than providing a merely detailed description, as is the case with the existing solutions. Moreover, machine-learning applications can be used for automatic tagging of publications to make content fully accessible. Such technologies are useful both in the production of alternative versions of non-accessible books and for born-accessible publications, in particular as far as backlist titles are concerned.

An area of application that touches both of the above mentioned categories and could help publishing houses is in the field of copyright management; here, technology innovation is required to streamline communication and access to rights information in the digital market, thus enabling a wider lawful access to digital content. A step further in this research area is to use artificial intelligence for automated recognition of textual and visual content on the web, and to return relevant licensing information to end users. Moreover, by combining artificial intelligence and Blockchain technologies it would be possible to explore the usage of intelligent agents to request license information for any digital content on the web and, when licensing conditions fit with users' requests, to enable automated micropayments. On a similar note, artificial intelligence could help online content sharing services ensure that unauthorised copyright works are not available on their services, without preventing the availability of non-infringing materials.

Another area of application – also on the side of the customers – is using AI to enhance the discoverability of publishing products from the byzantine amount of options, typically in the form of book recommendations. These are meant to better match books and readers. The concept can be based on the audio similarity computation systems used in the music industry. In general, ways are being sought to not limit analyses to mass-behaviour (what other people bought and viewed), focusing instead on other, content-related features (such as topic, ´suspense´, characters), deriving recommendations based on its analysis. As on top of social media (virtual ´word of mouth´), corresponding online sources become increasingly important in the e-retailing of books and in social reading, such algorithms might develop to be among the most relevant vehicles for reader choices.

*Challenges*

- Evaluate which applications are affordable and make business more effective and/or help to develop better products - and then be able to implement them
- Short to medium term self-interest: try to avoid pitfalls like ´AI silos´ (getting trapped in free AI application [APIs] by the large American platform companies and building your IT infrastructure around them) or to fall for dysfunctional ´AI´ badging
- Long term societal interest: take the special responsibility of the publishing industry seriously, consider the limits of the technology, educate readers about it (and its limits) - and most of all: help to set them free from filter bubbles and echo chambers
- Addressing intellectual property rights issues related to the input and output of AI
- More in general, texts and data are the fuel of any AI application. As far as AI is a very promising area for future business and society, the control of robots' fuel is an important political matter. If 'machine readable' refers to particular codes, who is in charge to define which codes are legally valid? Is it necessary a standard language, to avoid any lock-in to particular platforms' methods to communicate the same information?

### 2.4.2 Media Access Services

*Diffusion and consumption*

Technology is transforming the way we work, live and entertain ourselves. Yet, television (watched on a TV set or via the Internet) is still the preferred medium of Europeans: more than nine out of ten (96%) Europeans watch TV at least once a week. Europeans predominantly watch television on a TV set [3]. But television is changing. It is becoming more connected. Hybrid Broadcast Broadband TV (HbbTV) is an international, open standard for interactive TV, which enables innovative, Interactive services over broadcast and broadband networks [4]. How can the industry guarantee that as many people as possible benefit from this technological innovation?  And, if Europe is to become a world leader in accessibility, a topic raised recently by the European technology platform NEM (New European Media, 2016) [5], what steps are still needed?

The principle of subtitling for the deaf and hard of hearing was introduced in the United Kingdom in the early 1970s to meet the requirements of the hearing impaired people to access TV programs. This first system (Ceefax created in 1972 and Oracle), became widespread on the television channels in 1976. At that time (1976) appeared in France the Antiope system [1]. Teletext subtitling made its first appearance on Antenne 2 (France Televisions 2 today) on November 1st. 1983, on France 3 and TF1 in 1984, on Canal + in 1994, and on Arte in 1998. The Antiope system has been replaced by the European standard Ceefax on January 1st, 1995. Today, subtitling for the deaf and hard of hearing is governed by the European standard (EBU / EBU-N19).

Now it becomes necessary to define subtitling in a contextualized way, that is to say taking into account the current state-of-the-art and technique(s) (notably those based on AI in general, and deep learning in particular), the regulations and relevant market(s), describing it as:

- A sequence of subtitles that restores the meaning of the speaker's speech while adapting his/her words if necessary,
- Free of spelling errors and misinterpretations,
- Respectful of the standards in the country / countries concerned. These standards are techniques (objective standards), and artistic (subjective rules).

This definition effectively eliminates all automatic captioning providers that abound in the relevant market and do not meet any of the points in the above definition. The future technical challenge is to transform / improve this state of fact. Automatic translation out of AI and deep learning tools will allow to respond to the explosion of content (Big Data aspects), compliance with digital accessibility legislation, and the reduction of production costs.

The economic interest is threefold:

- First, the automation of the adapted subtitling chain will allow productivity gains that reduce unit costs and increase the volume of processed data.
- Then the production of multilingual subtitles will allow a wider commercialization of the audiovisual contents produced. Distributors of videos and audiovisual programs will be able to market their international programs more easily thanks to the presence of multilingual subtitles.
- Finally, the decrease in subtitle production costs will make captioning accessible to many new players for whom the cost makes captioning impossible.

Future short and mid term innovation trends include:

- For broadcasting: developing and improving sign language production, Audio Description for content (videos and books) with the facility to deliver dialogue and ambiance elements of the soundtrack separately, achieving robust subtitling performance across genres and increasing interoperability, allowing users to consume personalised automatic live subtitles anywhere.
- For web access developments: industrialize existing prototypes e.g.: subtitle renderer; inlay/screen overlay (incrustation) of sign language interpreter; advanced audio functions; improve the quality of automatically generated subtitles, reliable Speech-To-Text technologies, improve avatar based signing services, develop and integrate additional accessibility services into existing online platforms.

The market for subtitling is highly fragmented and operates primarily at the national levels. There are no comparative European nor international studies concerning the players involved, their market shares or their intentions for technological development.

According to one of the rare studies on dubbing needs and practices in the audiovisual industry in Europe, there are 631 dubbing and subtitling companies in 31 European countries, 160 of which are leading companies. Their overall turnover was estimated between 372 and 465 million Euros. 84 companies are located in France, Italy, the United Kingdom and Germany, accounting for 64% of the turnover on these activities. 30% of sales (turnover) would be made on audiovisual work.

At the time, as today, the circulation of programs and the transfer of language could be further developed, accessibility is slowly improving in view of the European directives and some dedicated projects that have emerged, but is not applied equally or consistently everywhere. The absence of multilingualism penalizes certain future technological innovation perspectives, and the quality of subtitles is not always present going together with increasing pressure on the translation professionals.

Legal obligations [2] have been a real opportunity for many actors. In France, the production of subtitles by its specialized service has increased from 6,045 hours at the end of 2008 (made entirely by authors / adapters) to 8,380 hours at the end of 2009 (with 1,939 hours of live programs) and 13 140 hours at the end of 2010 (with 5 097 hours performed live).

Thus, we observe that the legal context has completely shaped and redesigned the production mode (through the introduction of speech recognition software) and the routing of captioning on the air and the production of subtitles. The regulatory issue clearly favors market opportunities.

Finally, the subtitling market will certainly concern cinema and television, but it also concerns more and more advertising and the world of performing arts, education & training, or even the integration of foreigners into a country.

Many countries (France, EU, Australia, Canada, Germany, Hong Kong, India, Ireland, Italy, Japan, Netherlands, New Zealand, Norway, Spain, Great Britain, United States, etc.) have adopted legislation similar to French legislation (cited above) on digital accessibility resulting in the need for the production of appropriate subtitles for audiovisual content.

The leading countries in these areas include the United Kingdom and the United States of America. Extrapolating from the French market, we can reasonably estimate that the annual market for the production of subtitles adapted to audiovisual broadcasters at a global level of several hundreds of millions or even billions of Euros per year. For players in this market, at European level, the Ericsson Group, including its subsidiary Red Bee Media, is a major player in the field of accessibility, providing more than 200,000 hours of subtitling each year, including 80,000 of live subtitling. In the UK, BTI Studios has produced more than 350,000 hours of captioning each year. Finally in the United States,

3PlayMedia is a major player. Among French actors currently known are MFP, ST501, Blue Elements, Dubbing Brothers, Titra Films, Imagine, etc.

*Challenges*

- Fluidize/streamline the circulation of audiovisual (or video) programs through machine translation, while humans focus on the quality of work, for example.
- Machine translation would also make it easier for television channels to acquire new foreign customers and allow them to invest more easily in extra-European programs without investing too much in subtitling;
- Encourage more synergies and convergence between subtitling and the development of multilingualism or the integration of foreigners (migrants for example) in a given country.
- Develop AI tools for automatic translation to sign language, and from sign language to text
- Develop AI tools for robust automatic translation of subtitles (multi-languages)

*References*

[1] Acquisition Numérique et Télévisualisation d'Images Organisées en Pages d'Ecriture / Digital Acquisition and Televisualisation of Images Organized as written Pages

[2] in France:

https://www.legifrance.gouv.fr/affichTexte.do?cidTexte=JORFTEXT000000809647&categorieLien=id

[3] Media use in the European Union 2014

http://ec.europa.eu/public_opinion/archives/eb/eb82/eb82_media_en.pdf

[4] https://www.hbbtv.org/overview/#hbbtv-overview

[5] NEM-Access Report: Opening Doors to Universal Access to the Media. February 2016.

http://nem-initiative.org/wp-content/uploads/2016/03/NEM-ACCESS-Policy-suggestions.pdf

### 2.4.3 News

AI is gradually changing the news media business, impacting all steps from production to consumption.

*Creation and production*

On the production side, information gathering and synthesis is benefiting and will continue to benefit from increasing technological achievements to facilitate the analysis and cross-examination of heterogeneous information sources in multiple languages, including linked open data and crowdsourcing, to help validating information and facts on a large scale (so-called *fact checking*), to automatically provide insightful, potentially personalized, digests including enlightening visualization and summarization. Examination of the Panama Papers, leveraging natural language processing and text mining techniques in conjunction with database technology and graph visual analytics, is a recent example of this trend[19], which also points at the limitations of today's technology. The recently ended EU project YourDataStories focusing on linked data for investigation journalism is another meaningful example, however limited to homogeneous well-structured data. Addressing technology for heterogeneous sources, the Inria project lab iCODA focuses on the seamless

---

[19] http://data.blog.lemonde.fr/2016/04/08/panama-papers-un-defi-technique-pour-le-journalisme-de-donnees in French

integration and exploration of knowledge bases, public databases and curated content collections for data journalism.

Fact checking is another emblematic use-case where AI is bound to make a difference, as highlighted in recent initiatives such as the EU projects [Pheme](), [REVEAL]() or InVid. Making use of knowledge representation, natural language processing, information extraction, image retrieval and image forensics deeply modifies the debunking of fake news, while social network analytics provides the means to better understand how and by whom fake news are propagated so as to facilitate their dismantling. On the other hand, image and video manipulation is rapidly improving, in particular with recent advances in deep learning for text and image synthesis (cf. [fake discourse of President Obama presented at SIGGRAPH 2017]()), and fake news producers will sooner or later become aware of the methods used to track them and find workarounds. This calls for the development of efficient countermeasures and adversarial approaches.

*Diffusion, consumption*

On the consumption side, AI technology also modifies in depth our habits. User profiling and recommender systems are on the verge of being widely used as the number of information sources critically increases. This increase of sources also calls for mechanisms and general public tools for users to assess the reliability of the information they are provided with, beyond the traditional work of press agencies, and possibly across language barriers. News aggregation and summarization is also key in today's news consumption and still requires significant work on automatic multimodal summarization and story-telling easily adaptable to a user's personal expectations, on new content generation, etc. Last but not least, participative journalism is progressively becoming a standard (see, e.g., tweets embedded in newspaper articles or in news shows, videos and photos of events being taken by witnesses and incorporated in professional news reports). This growing trend, which gradually shifts journalist work from professional redaction to the general public, from official news providers to social networks, must be accompanied with intelligent processing tools to maintain high-quality information channels.

*Challenges*

- Heterogeneous data integration and querying, with ontology-based access (making the most of participative input, capitalizing on existing knowledge bases and public open data);
- Efficiency, trust and timeliness in information extraction and knowledge discovery (i.e., better collaborative, up-to-date and easy-to-maintain knowledge bases), as well as in content production, whether automatic or not;
- Improve the security of multimedia information retrieval systems and image/video forensics; better personalization and recommendation;
- Trust and transparency of algorithms, and potentially of information (blockchain?) are also at stake here.

### 2.4.4  Social Media

Social media, today mostly dominated by large companies in the US such as Google, Twitter, Instagram or Facebook, have become an important channel for information and entertainment, conveying huge amounts of personal information that can be used as a proxy to study and monitor people's mind set on a topic or on a product. AI technology has already revolutionized the way social media content is indexed, searched and used, with key technology such as object, face or action recognition in images and videos, entity detection in texts, or opinion mining and characterization. Highly distributed recommender systems exploiting user profiling are today also instrumental to social networks, including for ad placement. Beyond the analysis of user-generated content for indexing and search purposes, monitoring content and users on social networks can provide valuable information and knowledge on specific communities, on people's behaviour and opinions, on societal trends, etc.

To learn more, the interested reader can refer to a NEM position paper dedicated to social media, entitled "Towards the future social media".

## 3   Identifying generic technological and societal challenges

The previous section has proposed an overview of where and how AI is used in media and creative industries. It has also identified a list of specific technological/scientific challenges for each individual application area. Based on these, we can identify some more generic, application-independent technological challenges. Additionally, the media and creative industries products aim at being used by the society at large. Thus, using AI in this context does not only raise technological and scientific challenges, but also societal ones, notably including ethics.

### 3.1   Transversal technological challenges

**Data.** Many modern AI techniques, and Deep Learning in particular, require vast amount of training data to be calibrated, in order to be able to recognize patterns in these data, and then make decisions and predictions. A generic challenge is thus to acquire and build large data bases of relevant data and signals, for each application area. Moreover, such data need to be labeled with the true patterns of interest since most recent AI method are supervised. This poses a particular challenge for the creative industries as with many applications, the patterns to be detected are the users' preferences (choice of music or video preferences, cinema, news or video game content preferences, display format preferences, etc.). Such preferences are rarely explicitly provided by the users, nor asked to them. A generic challenge is thus to infer such preferences from the users' behaviour (history of choices, consumption time, etc) or various other indirect measures from the user, such as various physiological signals. Finally, acquiring and labelling large amount of data for media and creative applications is a challenge in itself. As some AI unsupervised or weakly-supervised techniques have started to do[20], a next challenge would thus be to design AI tools that can work from very little labelled data, or even from unlabelled ones.

**Robustness.** Many applications in the media and creative industries in which AI tools are used, use them for pattern recognition, e.g., to recognize speech, images, sounds, styles or user's mental states. While recent DL tools have greatly improved the recognition accuracies for several medias, notably images, videos or speech, current recognition performances are still not perfect for those, and still require substantial improvements for others. There is thus a need to design AI tools that can more robustly and more reliably recognize patterns in data, even when these data are noisy and/or non-stationary, which is the case for many data in the creative industry sector. Such tools should also be able to be robust even with missing data.

**Cross-domain methods**. Modern AI tools have proven particularly efficient in many domains, for analyzing various data and media. However, most of them work very well for a very specific domain only, with a very specific type of data only. There is thus a need for methods that can work across domains, or to develop methods that can make AI tools developed for specific domains to interact together. Such needed cross-domain AI tools notably include cross-modality analyses, to analyse and index at the same time text, images or speech for instance. This can notably prove very useful for analyzing or synthesizing news - whose content can be presented in various modalities - or for making content accessible, by translating or presenting it in speech, text or sign language for instance. Similarly, cross-modality could enable to perform media and content search from various modalities, e.g., to search images from a text or speech description. Cross-domain methods could also be used to perform multiscale analysis, e.g., by working on multiple temporal scales for music style analysis, or by working on multiple spatial scales, to analyze or synthesize large real or virtual environments. Finally, cross domain methods could enable more efficient style transfer (e.g., musical style transfer or image style transfer), even with non-realistic content, e.g., non-photorealistic paintings, or musical timber styles.

---

[20] AlphaZero: https://deepmind.com/blog/alphago-zero-learning-scratch/; Webvision: https://www.vision.ee.ethz.ch/webvision/

**Human-machine cooperation.** In the media and creative industries, the provided services and products often revolve around the users, who are creating, producing or consuming the content. When including AI tools in these products, we thus still need to consider user-centred designs. In particular, we should study and consider how the human users will understand and make use of such AI tools, in order to ensure human-machine cooperation. In other words, we want to make sure that the users will make the best use of these new AI tools, so this can increase their usability and their user experience (UX). This thus means that using AI tools in the creative industry sector is not only an AI problem, but also a Human-Computer Interaction (HCI) one. AI and machine learning experts should thus team-up with HCI, UX and design experts to come up with useful, usable and enjoyable products. For instance, this means that the tools provided to users should work efficiently and in real-time, should explain what they do to users rather than being full black boxes, and enable users to extend them, to adapt them, and to include their own constraints. It also means such AI tools should understand and adapt to users preferences.

**Accountability**. Recent deep learning methods have reached various successes in tasks such as recognizing speech, images, music tracks etc. But such methods are also known to be rather opaque regarding the criteria used to make predictions and take decisions. Performance accuracy has been a predominant measure of success for these methods but, recently, a good deal of research has started to work on their accountability. This is critical in the media and creative industry.

**Ethics-by-design.** Last but not least, all these AI tools can infer many information about the users' behaviours, preferences and needs. These tools can also guide or even influence the user's behaviours. There is thus a pressing need to consider ethical questions for all those systems (see also next section for more details). From a technological/scientific point of view, this means that a strong challenge is to come up with ethics-by-design algorithms and tools, that ensure they respect some ethical principles. This also means that we need algorithmic tools that are transparent and trustable.

## 3.2 Societal challenges

These disruptive technologies indeed raise new ethical and regulation challenges. These aspects are well addressed in several documents, like the Villani et al report on « AI for humanity »[21] and will not be rephrased here. Similarly the General Data Protection Regulation (GDPR)[22] aims primarily to give control to citizens and residents over their personal data and to simplify the regulation. Here we report societal challenges that are especially relevant regarding AI and media, some being seldom quoted in published texts, although essential.

---
[21] https://www.aiforhumanity.fr/pdfs/9782111457089_Rapport_Villani_accessible.pdf

**Automation.** AI is often seen as an efficient way to "automatise" increasingly complex tasks. However, seeing AI as an opaque autonomous system introduces huge biases. Firstly, it hides who decides. If an AI is designing someone's news feed, recommending songs "driven by" her taste, a company, or group of humans, is making profit underneath. This goes far beyond the advertising submitted to us. Secondly, some aspects of the sovereign power tends to be appointed to those who have the power on the data and algorithms, e.g., in education, where the companies building educational digital resources decide in a much deeper way that with books, what is going to be the learning curriculum. Regarding the media, this can have a dramatic impact on freedom of expression. Finally, opaque autonomous system makes difficult for human-users, citizens, to understand how it works. Understanding how the systems works allows us to realize that this is not magic, and to construct a representation of what can be done or not with it. Regarding computer science in general, we are now beyond the common wrong idea that we do not need to understand how it works but simply (obediently) use it. This also applies to AI principle and it is a major goal for science outreach and science popularization to produce resources for everyone on these topics. Furthermore, in addition to developing computational thinking, this includes developing creativity (and not only using tools as it), developing critical thinking: We must neither be technophobe nor technophile, but technocritic.

**Ownership.** The fact is Google, Facebook, Amazon and Microsoft GAFAMs are, to a certain extend, leading the developments in AI, including by sharing open software widely used in research such as Tensorflow or Malmö. Shall we attempt to do better than they do with their huge resources or build on what they share? The solution might be elsewhere. On the one hand, consider GAFAMs want or need to enhance somehow their image with regard to geopolitical issues. On the other hand, clearly state the frontier between what they can propose, and what is definitely the sovereign domain (e.g., education, regulation, or health care). In education the Class´Code[23] project is paradigmatic with respect to Google for Education[24] (GfE) will of leading computer science education in France: this initiative has defended its independence with respect to the "giant", and proposes a complete French and soon European common good, without refusing to collaborate with GfE, which supports peripheral actions. Because Class´Code gathers more than 70 partners in the related field, it offers an independent leadership on its targeted topic. This could be generalized to other domains.

**Accessibility.** As described in section 2, AI tools could prove very efficient in making media accessible to all, through, for example, automatic machine translation of text and speech, or automatic subtitling and sign language generation. This shows the need for a continuing emphasis on media accessibility, while recognizing that many strides have been taken in Europe so far. This is well achieved through education; standardisation and legislation based on sound academic and industry research and by the involvement of all members of the value chain, not forgetting the users. Another accessibility challenge where AI can have an important impact is with regard to migration. Migration should be placed under perspectives to promote multiple knowledge (of and about those who arrive), about cultural differences, and the economical sides: many refugees are coming to Europe, and they can impulse economic growth. AI tools could support, for example, real time translation tools aiming at empowering human contact. Refugees need to be connected to the country's language, but also to their own language.  Besides the above mentioned translation tools, media (particularly Public Services) can serve as Educational and Knowledge Diffusion Platform for all  - both migrants/immigrants and local population. Content creators, creative people, and storytellers could be "encouraged" to produce content related to immigrants (beyond news and reports based on emergency).

---

[22] https://en.wikipedia.org/wiki/General_Data_Protection_Regulation
[23] https://classcode.fr
[24] https://edu.google.com/intl/fr_be/

**Science of belief.** While "weak" AI, (or technical AI, as properly defined by (Ganascia, 2017)[25]) is a reality, and corresponds to the fact that « machines can realize tasks which would have been considered as intelligent if realized by a human » as defined by Minsky[26], with the intrinsic limit of being always specific to a narrow cognitive task (as formalized by the no-free lunch theorem[27]). What is called "strong" or "global" AI (i.e., that an intelligence with consciousness will (in a near future) emerge from a machine able to reproduce and improve itself) is neither true nor wrong, but relates to a belief. Very honorable people in the world believe that trees have a soul, and other people believe in strong AI, that will either eliminate the humanity or create a human paradise, for some of us (Ganascia, 2017).

# 4 Conclusion

In this document, we showed that AI technologies can have multiple uses for media and creative industries, at three main levels: at the creation level - creating new media content, at the production level - editing and processing content, and at the consumption level - using and interacting with the content.

AI tools at these three levels have proven and/or will prove to be relevant and useful for a wide range of media and creative applications, being used for sound and music, images and videos, video games, design, engineering, news, social network or accessibility, among possibly many others. For instance, AI tools have been or could be used to provide personalized media content to consumers, using recommender systems or by modelling these consumers, in order to propose them automatically music, movies, news or games that they are likely to enjoy, or by even dynamically adapting the content providing to them to maximize their user experience. This also includes adapting the content and its format to possible disabilities, to make accessible and inclusive media. AI tools could also be used to clean, enhance or edit media content automatically, e.g., by transferring styles across musics, images or videos, among others. It could also be used to transfer information and semantics across media modalities, or to search information across modalities, e.g., searching images or videos from sound and text, or the other way around. Finally, AI tools could be used to create and synthesize new contents from large data bases of existing content, e.g., with automatic news extractions and creation, or could be used to help and assist artists and designers in creating new contents faster and more effectively, e.g., by automatically creating 3D objects from quick 2D sketches.

All these potentials usages and applications come with a number of challenges to be solved, both at the technological and societal levels. This includes producing, gathering and/or labelling the large amount of data that can be needed by AI methods such as Deep learning. This also includes designing methods that are robust to noise and non-stationarity in the medium used as input (sound, images, text, etc.), and that can work across modalities and/or applications. Users, whether using the AI tools or being impacted by them, should also not be forgotten. There is a need for AI tools that can adapt to the users, consider their needs and constraints, and that the users can adapt, i.e., to work as a partner to humans rather than as black boxes. AI methods should also become accountable, to avoid possible biases, as well as ethics by design, to ensure that choices made by or with the help of an AI are necessarily ethic. This naturally brings us to societal challenges of the use of AI for media and creative industries: notably ensuring unbiased, ethic and empowering application and use of AI in media, the need for strong European tools and players in this field, or the possible benefits of AI in media for accessibility and migrations.

---

[25] http://www.seuil.com/ouvrage/le-mythe-de-la-singularite-jean-gabriel-ganascia/9782021309997
[26] http://binaire.blog.lemonde.fr/2016/01/29/lintelligence-artificielle-debraillee/
[27] https://en.wikipedia.org/wiki/No_free_lunch_theorem

Altogether, R&D&I in AI for media and creative industries is a rich, relevant and very promising endeavour, with already a number of recent innovations, and arguably even more expected innovations and products, if we can solve the challenges mentioned above. Moreover, this report probably identified only parts of the applications, areas and challenges related to AI for media and creative industries. Additional ideas and R&D&I directions are certainly also promising and worth exploring, given the diversity of tools, needs and applications in this field. Further creativity in the use of AI for media should thus also be encouraged.